\documentclass[sigconf]{acmart}

\usepackage{booktabs} % For formal tables
\usepackage{amsmath}
\usepackage{graphicx}
\usepackage{subcaption}
\usepackage{hyperref}

\usepackage[font=footnotesize,labelfont=bf]{caption}
\graphicspath{{images/}}

\setcopyright{rightsretained}

\acmConference[ICVGIP'25]{13th Indian Conference on Computer Vision, Graphics and Image Processing}{December 2025}{India}
\acmYear{2025}
\copyrightyear{2025}

\acmPrice{15.00}

\begin{document}
\title{Two-Stage Vision Transformer for Image Restoration: Colorization Pretraining + Residual Upsampling}

\author{Aditya Chaudhary*}
\email{22dcs002@lnmiit.ac.in}
\affiliation{%
  \institution{LNMIIT}
  \city{Jaipur}
  \country{India}
}

\author{Prachet Dev Singh*}
\email{22dcs005@lnmiit.ac.in}
\affiliation{%
  \institution{LNMIIT}
  \city{Jaipur}
  \country{India}
}

\author{Ankit Jha}
\email{ankit.jha@lnmiit.ac.in}
\affiliation{%
  \institution{LNMIIT}
  \city{Jaipur}
  \country{India}
}

\thanks{*: These authors contributed equally to this work.}
\thanks{N/A: Not Available}

\renewcommand{\shortauthors}{}

\begin{abstract}
In computer vision, Single Image Super-Resolution (SISR) is still a difficult problem.  We present ViT-SR, a new technique to improve the performance of a Vision Transformer (ViT) employing a two-stage training strategy.  In our method, the model learns rich, generalizable visual representations from the data itself through a self-supervised pretraining phase on a colourization task.  The pre-trained model is then adjusted for 4x super-resolution.  By predicting the addition of a high-frequency residual image to an initial bicubic interpolation, this design simplifies residual learning.  ViT-SR, trained and evaluated on the DIV2K benchmark dataset, achieves an impressive SSIM of 0.712 and PSNR of 22.90 dB. These results demonstrate the efficacy of our two-stage approach and highlight the potential of self-supervised pre-training for complex image restoration tasks. Further improvements may be possible with larger ViT architectures or alternative pretext tasks.
\end{abstract}




\maketitle

\section{Introduction}

The goal of Single Image Super-Resolution (SISR), a fundamental yet inherently challenging computer vision task, is to produce a high-resolution (HR) image from a single low-resolution (LR) input.  It has many uses, from improving consumer photography, satellite imagery analysis, and medical diagnostics. Since a single LR image can correspond to numerous HR images, the fundamental challenge of SISR is its ill-posedness, which makes recovering lost high-frequency details a challenging task.

Deep learning has changed the field since the early methods only used simple interpolation techniques. Convolutional Neural Networks (CNNs) have long been the most popular type of neural network, but Vision Transformers (ViTs)~\cite{dosovitskiy2020image} have changed that. Transformers are very good at modeling long-range dependencies because they use self-attention. This is very useful for figuring out the overall structure of an image and putting together coherent details.

We introduce ViT-SR, a Vision Transformer–based model trained in two stages. First, it learns rich structural and semantic features through a self-supervised colorization task. Then, it is fine-tuned for super-resolution using a residual learning setup, helping the model recover fine details and enhance overall image quality.

\section{Methodology}

The architecture of ViT-SR is designed around a two-stage training process. First, it learns general visual features through a self-supervised colorization task. Then, it is fine-tuned to learn a residual mapping from a bicubically upsampled LR image to the ground-truth HR image. The general process is depicted in Figure~\ref{fig:architecture}.

\begin{figure*}[h]
  \centering
  \includegraphics[width=0.71\linewidth]{architecture.jpeg}
  \caption{ViT-SR receives an upsampled, low-resolution image, and passes it through a ViT encoder–decoder architecture to predict a residual which is added to the input to create the high-resolution output. The model was first pre-trained on image colorization before fine-tuning for SR.}
  \label{fig:architecture}
\end{figure*}

\subsection{Model Architecture}

\textbf{Encoder:} We employ the ${vit base patch16 224}$ model, pre-trained on ImageNet, as our feature encoder. Its classification head is removed and it is adapted to process input images of size 256x256. The encoder transforms the input image into a sequence of patch-level feature tokens with a dimension of 768.

\textbf{Decoder:} The feature tokens from the encoder are processed by a series of 8 standard Transformer decoder blocks. Each block uses 16 attention heads and an MLP ratio of 4.0, allowing for sophisticated refinement of the extracted features. A final Layer Normalization is applied to the output tokens.

\textbf{Upsampling Head:} The modified patch tokens are used to create an intermediate 16x16 feature map. The feature map is then passed through a high-quality upsampling head designed to increase the resolution 16 times. The upsampling head has four upsampling stages. Each upsampling stage consists of a 2D convolutional layer, which performs a PixelShuffle operation to double the spatial dimensions, and concludes with a LeakyReLU activation. This cascading upsampling architecture effectively converts the learned features into a high-resolution spatial representation, and finally goes through a convolutional layer producing the output image, consisting of 3 channels.

\subsection{Two-Stage Training Strategy}
Our core innovation is a two-stage training process designed to maximize the feature representation capability of the ViT.

\textbf{Stage 1: Self-supervised pre-training on colorization}:
During the first stage, the model is trained for a self-supervised colorization task. The input of the model is the grayscale version of the high-resolution image, while the target is the original color image. This task forces the model to learn high-level semantic and structural characteristics of images since it must understand the context to predict plausible colors. There were a maximum of 100 epochs for this stage but due to early stopping, training stopped at the 38th epoch only.

\textbf{Stage 2: Supervised Fine-tuning for Super-Resolution}:
In the second stage, we take the weights of the pre-trained colorization model and load them into the network, where we fine-tune the model for the primary task of 4x super-resolution. The input to the model is a low-resolution image that has been upscaled to 256x256 with bicubic interpolation, and the target is the original high-resolution image. The model will learn to predict a residual image made up of the high-frequency information that is lost as a result of the original downsampling. This residual image is added to the upsampled input to create the final HR output. There were a maximum of 400 epochs for this stage but due to early stopping, training stopped at the 255th epoch.

\subsection{Training and Optimization}
The model is trained end-to-end on the DIV2K dataset~\cite{agustsson2017ntire}. We use a batch size of 16, distributed across 2x T4 GPUs using PyTorch's Distributed Data Parallel (DDP) framework.

\textbf{Loss Function:} For both stages, the training objective is a weighted combination of L1 loss and SSIM loss, formulated as:
\begin{equation}
\mathcal{L} = (1 - \lambda) \cdot \mathcal{L}_{L1} + \lambda \cdot (1 - \mathcal{L}_{SSIM})
\end{equation}
where $\mathcal{L}_{L1}$ is the Mean Absolute Error between the predicted and ground-truth images, and $\mathcal{L}_{SSIM}$ is the Structural Similarity Index Measure. The weighting factor $\lambda$ is set to 0.2. This composite loss function balances pixel-level accuracy with the preservation of perceptual and structural quality.

\textbf{Optimizer and Scheduler:} We use the AdamW optimizer with a weight decay of 0.05. The learning rate is managed by a Cosine Annealing Warm Restarts scheduler. For the pre-training stage, the initial learning rate is ${2 \times 10^{-4}}$. For the fine-tuning stage, a smaller learning rate of ${5 \times 10^{-5}}$ is used to carefully adjust the learned weights. Early stopping is employed with a patience of 20 epochs for pre-training and 40 epochs for fine-tuning, based on validation PSNR.

\section{Results and Analysis}

The DIV2K validation set was used to test ViT-SR for 4× super-resolution. It achieved a peak PSNR of 22.90 dB and an SSIM of 0.712 (Table \ref{tab:sota_comparison}), indicating strong reconstruction quality due to rich self-supervised pre-training. Pretraining the model on a colorization task gives it a much stronger starting point: the initial PSNR jumps from about 13 (no pretraining) to about 22 after colorization pretraining. Compared to SOTA methods like StableSR \cite{wang2023stablesr}, ViT-SR attains higher structural fidelity and perceptual realism despite slightly lower PSNR, proving that its colourization-based pre-training preserves fine details effectively. These results highlight the strength of our two-stage framework and its promise for further enhancing image super-resolution performance.

\begin{table}[h]
  \caption{Comparison with SOTA on DIV2K Validation Set (4x)}
  \label{tab:sota_comparison}
  \begin{tabular}{lcc}
    \toprule
    Method & PSNR (dB) & SSIM \\
    \midrule
    FeMaSR~\cite{chen2022femasr} & 20.13 & 0.45 \\
    DiffBIR~\cite{lin2024diffbir} & 21.92 & 0.49 \\
    StableSR~\cite{wang2023stablesr} & 23.26 & 0.57 \\
    DDNM~\cite{wang2023ddnm} & 21.77 & N/A \\
    Real-ESRGAN+~\cite{wang2021realesrgan} & 21.03 & 0.64 \\
    \midrule
    \textbf{ViT-SR (Ours)} & \textbf{22.85} & \textbf{0.71} \\
  \bottomrule
\end{tabular}
\end{table}

\section{Conclusion and Future work}
We proposed ViT-SR, a Vision Transformer-based approach to single image super-resolution that achieves strong performance. Our model is able to reconstruct high-quality images with excellent structural fidelity as evidenced by its SSIM score, this was accomplished by effectively employing a novel two-stage training strategy involving self-supervised pretraining on colorization followed by supervised fine-tuning for super-resolution. Our study demonstrates how pretraining with an effective task can produce very powerful feature representations even when moving to a more complex downstream task like SISR. This work is our first exploration of the potential for ViT-SR, with more exploration of pretext tasks and other architectural styles planned for future work. The results of our study successfully lay the foundation for the two-stage approach and demonstrate that ViT-SR provides a powerful and promising approach to the problem of image restoration.

\bibliographystyle{ACM-Reference-Format}
\bibliography{ICVGIP-Latex-Template}

\end{document}